\documentclass[letterpaper, 10 pt, conference]{ieeeconf}
\usepackage{amsmath,amsfonts}
\usepackage{array}
\usepackage{booktabs}
\usepackage{array}
\usepackage{colortbl}
\usepackage{enumerate}
\usepackage[utf8]{inputenc}
\usepackage[T1]{fontenc}
\usepackage{amsfonts}
\usepackage{amssymb}
\usepackage{tabularx}
\makeatletter
\let\NAT@parse\undefined
\makeatother
\usepackage{hyperref}
\hypersetup{
    colorlinks=true,
    linkcolor=blue,
    filecolor=magenta,      
    urlcolor=cyan,
    }
\IEEEoverridecommandlockouts 
\usepackage{algorithm}  
\usepackage{algpseudocode}  
\usepackage{amsmath}  
\usepackage{breqn}
\usepackage[caption=false]{subfig}
\usepackage{textcomp}
\usepackage{stfloats}
\usepackage{url}
\usepackage{verbatim}
\usepackage{graphicx}
\usepackage{cite}
\usepackage{color}
\usepackage{colortbl}
\usepackage[most]{tcolorbox}
\let\labelindent\relax
\usepackage{enumitem}
\usepackage{multirow}
\usepackage{diagbox}
\usepackage{booktabs}
\usepackage{xcolor}
\usepackage{listings}
\definecolor{lightblue}{rgb}{0.68, 0.85, 0.90}
\definecolor{lightpurple}{rgb}{0.87, 0.81, 0.95}
\definecolor{lightlightgray}{rgb}{0.95,0.95,0.95}
\definecolor{lightpink}{rgb}{1.0, 0.87, 0.87}
\definecolor{darkgreen}{rgb}{0.0, 0.5, 0.0}
\definecolor{darkblue}{rgb}{0.0, 0.0, 0.5}
\definecolor{orange}{rgb}{1.0, 0.5, 0.0}
\definecolor{purple}{rgb}{0.5, 0.0, 0.5}
\definecolor{darkred}{rgb}{0.6, 0.0, 0.0}

\title{\LARGE \bf PrefMMT: Modeling Human Preferences in Preference-based Reinforcement Learning with Multimodal Transformers}

\author{Dezhong Zhao$^{1,2}\dag$, Ruiqi Wang$^{2}\dag$, Dayoon Suh$^{2}$, Taehyeon Kim$^{2}$, \\ Ziqin Yuan$^{2}$, Byung-Cheol Min$^{2}$, and Guohua Chen$^{1}$
\thanks{$^\dag$These authors contributed equally.}
\thanks{$^{1}$College of Mechanical and Electrical Engineering, Beijing University of Chemical Technology, Beijing, China. \tt\small{[DZ\_Zhao, chengh]@buct.edu.cn}.}
\thanks{$^{2}$SMART Laboratory, Department of Computer and Information Technology, Purdue University, West Lafayette, IN, USA. {\tt\small{[wang5357, suh65, kim4435, minb]@purdue.edu}.}}
}

\begin{document}

\setlength{\abovedisplayskip}{0.5pt} 
\setlength{\belowdisplayskip}{0.5pt} 

\maketitle

\begin{abstract}
Preference-based reinforcement learning (PbRL) shows promise in aligning robot behaviors with human preferences, but its success depends heavily on the accurate modeling of human preferences through reward models. Most methods adopt Markovian assumptions for preference modeling (PM), which overlook the temporal dependencies within robot behavior trajectories that impact human evaluations. While recent works have utilized sequence modeling to mitigate this by learning sequential non-Markovian rewards, they ignore the multimodal nature of robot trajectories, which consist of elements from two distinctive modalities: state and action. As a result, they often struggle to capture the complex interplay between these modalities that significantly shapes human preferences. In this paper, we propose a multimodal sequence modeling approach for PM by disentangling state and action modalities. We introduce a multimodal transformer network, named PrefMMT, which hierarchically leverages intra-modal temporal dependencies and inter-modal state-action interactions to capture complex preference patterns. Our experimental results demonstrate that PrefMMT consistently outperforms state-of-the-art PM and direct preference-based policy learning baselines on locomotion tasks from the D4RL benchmark and manipulation tasks from the MetaWorld benchmark. Source code and supplementary information are available at \url{https://sites.google.com/view/prefmmt}. 
\end{abstract}

\section{Introduction}
Reinforcement learning (RL) has demonstrated significant prowess in robotics, enabling robots to acquire complex behaviors through trial and error \cite{kober2013reinforcement}. Despite its success, a major challenge in RL lies in designing appropriate reward functions, particularly in nuanced human-robot interaction scenarios \cite{wang2022feedback,munzer2017preference} and long-horizon tasks \cite{wang2023initial, wang2024initial}. Furthermore, issues such as reward exploitation can arise, leading to unintended and potentially hazardous robot behaviors \cite{yuan2019novel}.

Preference-based reinforcement learning (PbRL) \cite{lee2021b} has recently emerged as a promising approach to address these challenges by mitigating the complexities associated with explicit reward engineering. PbRL seeks to derive a preference-aligned reward model from human comparative feedback on pairs of robot trajectories, which is then used to optimize robot policies through vanilla RL. By integrating human feedback into the learning process, PbRL holds the potential to generate more desirable and aligned robot behaviors \cite{park2022surf}. However, this potential is highly dependent on the effective encoding of human preferences into the reward model. This process, known as preference modeling (PM) \cite{wirth2017survey}, is a non-trivial task that involves reconstructing the underlying reward structures implicitly reflected in human preferences.

\begin{figure*}[!t]
\centering
\includegraphics[width=\linewidth]{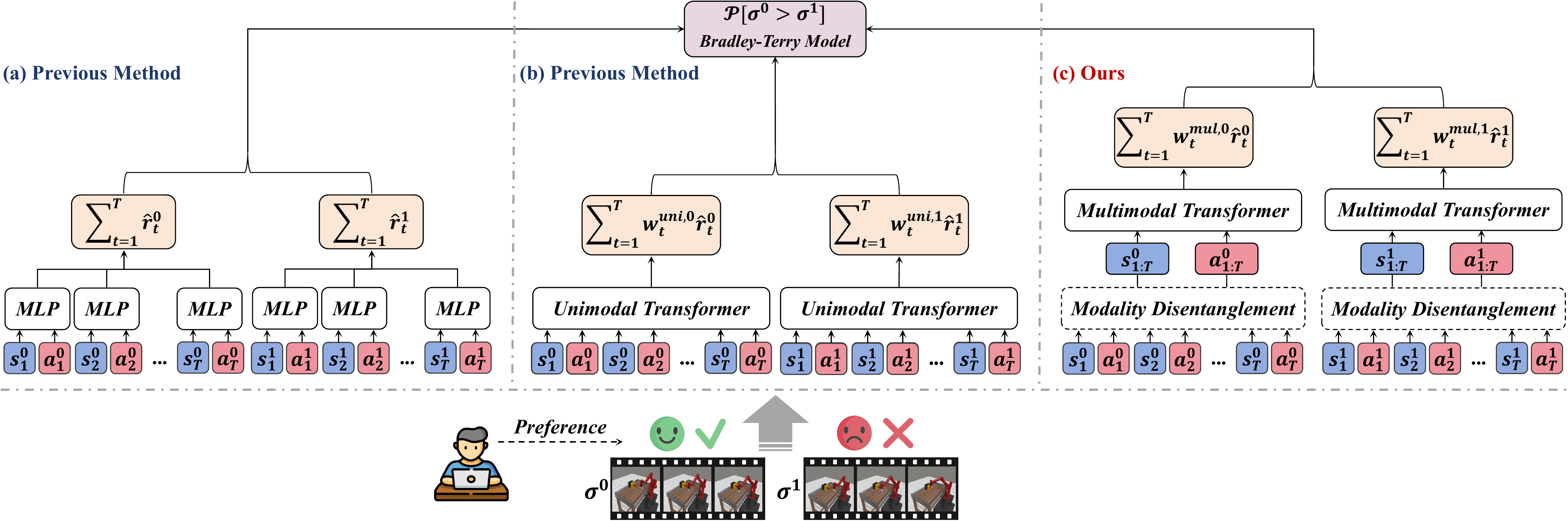}
\vspace{-15pt}
\caption{Comparison of previous methods and our approach (PrefMMT) for preference modeling in PbRL. \textbf{(a) Markovian Reward Modeling:} Assumes that human preference for a trajectory $\sigma$ is based on the equal sum of individual evaluations at each time step. The goal is to learn a Markovian reward model that assigns rewards based solely on the immediate state-action pair. (\textbf{b) Unimodal Sequence Modeling:} Regards a trajectory as a sequence and learns a series of non-Markovian rewards that depend on all previously visited time steps. By learning unimodal attention weights $w^{uni}$ with unimodal transformer networks, this method aims to infer temporal dependencies within the trajectory and identify critical time steps that significantly influence human judgments. \textbf{(c) Our Multimodal Sequence Modeling:} Recognizes the multimodal nature of a trajectory, disentangling the state and action modalities. By learning multimodal attention weights $w^{mul}$ via a multimodal transformer architecture, our approach captures both temporal intra-modal dependencies and inter-modal interactions between states and actions within the trajectory, leading to more nuanced credit assignment for human preferences.}
\vspace{-15pt}
\label{fig:comparision}
\end{figure*}

Most PbRL methods \cite{christiano2017deep,lee2021pebble,lee2021b,park2021surf,hiranaka2023primitive,metcalf2023sample,liu2022task,liu2022meta,wang2024prefclm} assume that human preferences for a robot trajectory, which consists of state-action pairs across multiple time steps, are an equal aggregate of independent evaluations of immediate time steps. Consequently, as shown in Fig. \ref{fig:comparision}(a), their aim of PM becomes learning a Markovian reward model, typically a multi-layer perception network, that generates preference rewards based solely on the instantaneous state-action pair. This Markovian assumption neglects the temporal dependencies that often govern human judgments. Human evaluations are typically influenced by earlier states and critical moments within the trajectory, where certain events or transitions carry greater weight \cite{bacchus1996rewarding,brafman2019regular}. Recognizing this limitation, recent research efforts \cite{early2022non,kim2023preference} begin to regard the PM as a sequence modeling problem \cite{shi2021neural}. As depicted in Fig. \ref{fig:comparision}(b), they view the trajectory as a sentence and learn a series of non-Markovian rewards, each depending on all previously visited time steps, using sequential networks such as long short-term memory and transformers. This approach enables the capture of temporal dependencies and the inference of critical events in the trajectory.

However, a significant limitation remains: existing methods typically treat robot trajectories as unimodal sequences, as seen in traditional sequence modeling \cite{vaswani2017attention}. In contrast, robot behavior sequence is inherently multimodal, involving both state and action modalities, each with its own dynamics \cite{wang2023trajectory}. Human judgments rely not only on the intra-modal temporal dynamics but also on the latent interactions between these modalities \cite{prasad2022mild,wang2021predicting,azar2024general}. This becomes especially important in complex real-world tasks where these interactions are more intricate and context-dependent. For instance, in manipulation tasks, humans may focus on the alignment between the object’s orientation and the gripper’s approach angle, or the relationship between the applied gripper force and the resulting object deformation, rather than solely on the temporal progression of individual state or action elements.

To bridge this gap, we introduce PrefMMT, a multimodal transformer network for modeling human preferences in PbRL. As illustrated in Fig. \ref{fig:comparision}(c), we advocate for the disentanglement of the state and action modalities from the trajectory, framing preference modeling as a multimodal sequence modeling problem. We present intra-modal and inter-modal dependency learning modules to hierarchically capture the complex relationships within and between modalities. This hierarchical approach enables PrefMMT to capture not only temporal intra-modal dependencies, inferring important state transitions or action sequences for human judgments, but also inter-modal interactions, such as how actions are conditioned on states, how states evolve in response to actions, and how these interactions contribute to overall trajectory preference. By explicitly modeling the multimodal nature of robot trajectories, PrefMMT aims to capture the nuanced dynamics of human preferences more comprehensively and efficiently. We summarize our key contributions as follow:
\begin{itemize}[leftmargin=*]
    \item We present PrefMMT as an efficient method to model human preferences in PbRL that explicitly accounts for the multimodal nature of robot trajectories.
    \item We develop a hierarchical multimodal transformer architecture that efficiently captures intra-modal dynamics within state and action modalities while modeling their inter-modal interactions, allowing for more nuanced and context-aware credit assignment in preference modeling.
    \item Through extensive experiments on locomotion and manipulation tasks from the D4RL \cite{fu2020d4rl} and MetaWorld \cite{yu2020meta} benchmarks, we demonstrate that PrefMMT models realistic human preferences more efficiently than state-of-the-art PM and direct preference-based policy learning methods, particularly in complex task scenarios.
\end{itemize}

\section{Background and Preliminary}
\label{PF}
\subsection{Preference-based RL and Preference Modeling}
Preference-based RL has emerged as a promising approach to address the challenges associated with reward engineering in traditional RL frameworks by incorporating human preferences into robot learning,  \cite{wang2022feedback,xue2024reinforcement,lee2021pebble,wang2024personalization}. The core of PbRL lies in modeling human preferences through a preference reward model, $\hat{R}_{\psi}$, typically implemented as a neural network with parameters $\psi$. This model is then used to train a robot policy via standard RL algorithms. 

Let $\sigma$ denote a trajectory of robot behaviors that consists of continuous state-action pairs across $T$ time steps, i.e., $\sigma = \{(\mathbf{s}_1, \mathbf{a}_1), \ldots, (\mathbf{s}_T, \mathbf{a}_T)\}$. A human evaluator provides preference labels $\Lambda \in \{0, 0.5, 1\}$ for pairs of trajectories. In this labeling scheme, "0" indicates a preference for trajectory 0 over trajectory 1, "1" indicates a preference for trajectory 1 over trajectory 0, and "0.5" indicates that the user is equally satisfied with both trajectories. By collecting multiple rounds of evaluations, we can obtain a preference dataset represented as: $\mathcal{D}_{\mathrm{p}}=\left\{\left(\sigma_i^0, \sigma_i^1, \Lambda_i\right)\right\}_{i=1}^{\left|\mathcal{D}_{\mathrm{p}}\right|}$. 

To learn the reward model $\hat{R}$ from the dataset, a Bradley-Terry model is utilized to estimate the preference likelihood of the reward model. For example, the likelihood that trajectory $\sigma^1$ is preferred over trajectory $\sigma^0$ is calculated as:
\begin{equation}
\mathcal{P}_\psi\left[\sigma^1 \succ \sigma^0\right]=\frac{\exp \left(\rho({\sigma^1};\psi)\right)}{\sum_{j \in\{0,1\}} \exp \left(\rho({\sigma^j};\psi)\right)}
\label{BT}
\end{equation}
where, $\rho(\sigma^j; \psi)$ denotes the overall reward output of the $\hat{R}_{\psi}$ for trajectory $\sigma^j$.

Then the reward model $\widehat{R}_\psi$ is trained by minimizing a cross-entropy loss function between the actual preference labels $\Lambda$ and the predicted preference probabilities:
\begin{equation}
\begin{split}
\mathcal{L}_{\psi}=-\sum_{\left(\sigma^{0}, \sigma^{1}, \Lambda \right) \in \mathcal{D}_{\mathrm{p}}} &(1-\Lambda) \log \mathcal{P}_{\psi}\left[\sigma^{1} \succ \sigma^{0}\right]+ \\ &\Lambda \log \mathcal{P}_{\psi}\left[\sigma^{0} \succ \sigma^{1}\right]
\end{split}
\label{loss}
\end{equation}

Most PbRL methods operate under the Markovian assumption, where the overall reward output is computed as the equal sum of rewards at each time step:
\begin{equation}
\rho_{\mathrm{MR}}(\sigma ; \psi)={\sum\nolimits_{t}} \widehat{r}_\psi\left(\mathbf{s}_t, \mathbf{a}_t\right)
\label{MR}
\end{equation}

However, human judgments are not just influenced instantaneous state-action pairs; they are also shaped by the contextual information provided by all previously visited states and critical moments in the trajectory \cite{bacchus1996rewarding,brafman2019regular}. Consequently, this Markovian assumption may not adequately capture the complexity of human preferences. Addressing this limitation, recent research \cite{early2022non,kim2023preference} has started to treat preference modeling as a sequence modeling problem, where a sequence of rewards is generated in response to the entire trajectory. These rewards are non-Markovian, meaning they depend on all previously encountered states and actions, and are also weighted to emphasize critical events in the trajectory:
\begin{equation}
\rho_{\mathrm{NMR}}(\sigma ; \psi)={\sum\nolimits_{t}} w_{t;\psi} \cdot \widehat{r}_\psi\left(\mathbf{s}_t, \mathbf{a}_t\right)
\label{NMR}
\end{equation}

While these methods can model the temporal dependencies of human evaluations for time-sequenced robot trajectories, they overlook the multimodal nature of a trajectory, which includes both state and action modalities. Consequently, they may fail to fully capture the latent interactions between states and actions that are crucial to human judgments \cite{prasad2022mild,wang2021predicting,azar2024general}. Our work addresses this limitation by employing multimodal sequence modeling with a hierarchical transformer network, enabling more nuanced preference assignment.

On the other hand, a recent line of research explores bypassing explicit PM by learning policies directly from preference data \cite{rafailov2024direct, hejna2023inverse, an2023direct, hejna2024contrastive}. However, these methods may encounter challenges related to sample efficiency and stability, as highlighted by \cite{nika2024reward, xudpo}. As a result, PM remains a critical research area due to its ability to capture nuanced human preferences and provide more interpretable reward structures with better generalization potential. Empirical results comparing our method to a state-of-the-art direct policy method \cite{hejna2024contrastive} demonstrate the potential to reinforce the value of PM by achieving higher efficiency and accuracy in human preference modeling.

\subsection{Transformers in Deep RL}
Transformers, originally developed for natural language processing tasks, have proven highly effective in sequence modeling \cite{vaswani2017attention,wang2024husformer} and have recently gained attention in deep reinforcement learning (RL) for their ability to capture long-range dependencies and process sequential data \cite{chen2021decision,hu2024transforming}. The Preference Transformer \cite{kim2023preference} represents the first attempt to apply transformers to PM by leveraging a unimodal, casual transformer architecture to capture temporal dependencies in robot trajectories. Building on this, our work takes a step forward by introducing a hierarchical multimodal transformer that not only captures temporal dependencies within state and action modalities but also models the complex inter-modal interactions that play a critical role in human preference evaluations.



\section{Methodology}

\subsection{Overview}
\begin{figure}[t]
\centering
\includegraphics[width=0.83\columnwidth]{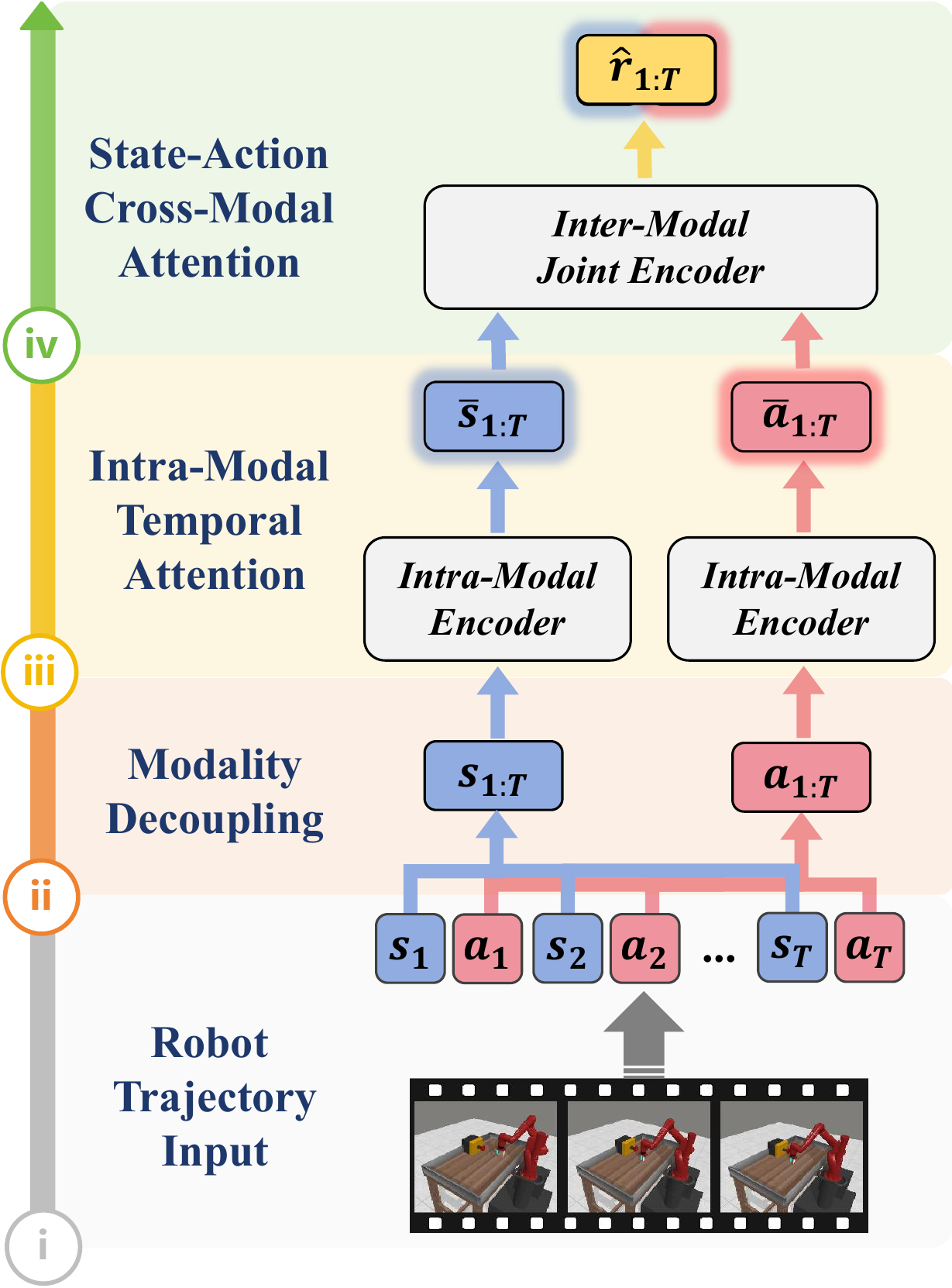}
\vspace{-5pt}
\caption{Illustration of the PrefMMT framework. Given a robot behavior trajectory as input, we first decouple the state and action modalities. Each unimodal sequence is then processed through an intra-modal encoder, where the temporal dependencies within the transitions of states and actions are explored. Subsequently, an inter-modal joint encoder captures the interactions between actions and states, outputting a series of non-Markovian rewards. }
\vspace{-15pt}
\label{fig:framework}
\end{figure}

In this section, we introduce PrefMMT, a multimodal transformer network designed to model human preferences over robot trajectories in PbRL. As illustrated in Fig. \ref{fig:framework}, PrefMMT operates in a hierarchical manner, first identifying critical temporal patterns within individual state or action modalities, which are decoupled from the robot trajectory, and then capturing the intricate interplay between these modalities. This dual-focus approach enables PrefMMT to identify key state transitions and action sequences that influence human judgments, as well as to model the reciprocal relationships between states and actions. Consequently, the framework elucidates how these multifaceted interactions collectively shape overall trajectory preferences, providing a nuanced interpretation of human preferences. In the following sub-sections, we introduce each module in detail.


\subsection{Modality Decoupling and Pre-processing}
In sequence modeling formulations \cite{chen2021decision,kim2023preference}, a robot trajectory $\sigma$ can be viewed as a sequence of length $T$: $\sigma = (\mathbf{s}_1, \mathbf{a}_1, \ldots, \mathbf{s}_T, \mathbf{a}_T)$. Given the multimodal nature of the trajectory, we first decouple the state and action modalities, forming state sequence $\mathcal{S} = {\mathbf{s}_{1:T}}$ and action sequence $\mathcal{A} = {\mathbf{a}_{1:T}}$, respectively.

Subsequently, following the data pre-processing procedures outlined in \cite{chen2021decision}, we pass the raw state and action sequences through embedding layers, denoted as $f_{e}$, to convert them into tokens of the same dimension, and then add time embeddings, presented as $\mathit{E}_{1:T}$, to incorporate temporal awareness within the trajectory:
\begin{equation}
\begin{aligned}
\mathbf{x}^{\mathcal{S}} &= f_{e}^{\mathcal{S}}(\mathbf{s}_{1:T}) + \mathit{E}_{1:T} \\
\mathbf{x}^{\mathcal{A}} &= f_{e}^{\mathcal{A}}(\mathbf{a}_{1:T}) + \mathit{E}_{1:T}
\end{aligned}
\end{equation}

\subsection{Intra-Modal Encoder}
The processed state and action sequences are separately fed into intra-modal encoders to capture intra-modal temporal attention, resulting in reinforced representations as:
\begin{equation}
\label{intra}
\begin{aligned}
\overline{\mathbf{x}}^{\mathcal{S}} &= \overline{\mathbf{s}}_{1:T} = f_{intra}^{\mathcal{S}}(\mathbf{x}^{\mathcal{S}}) \\
\overline{\mathbf{x}}^{\mathcal{A}} &= \overline{\mathbf{a}}_{1:T} = f_{intra}^{\mathcal{A}}(\mathbf{x}^{\mathcal{A}})
\end{aligned}
\end{equation}

Each intra-modal encoder, $f_{intra}$, is a 3-layer transformer network \cite{vaswani2017attention} with layer normalization and causally masked self-attention \cite{radford2018improving}, which prevents information leakage from future time steps in the sequence modeling setting. This step enables PrefMMT to capture modality-specific temporal dependencies that influence human judgments, such as precise transitions in the position of a target object in the state space or efficient and smooth changes in robot arm movements within the action space.

\subsection{Inter-Modal Joint Encoder}
The next step in PrefMMT is to pass the enhanced state and action sequence representations to an inter-modal joint encoder $f_{inter}$ to capture state-action cross-modal preference attention. This joint encoder also utilizes the causal transformer architecture as in $f_{intra}$, but replaces casual self-attention with casual cross-attention. We define Queries, Keys, and Values for state and action modalities as:
\vspace{+1pt}
\begin{equation}
\begin{aligned}
Q^{\mathcal{S}}, K^{\mathcal{S}}, V^{\mathcal{S}} &= W_q^{\mathcal{S}} \cdot \overline{\mathbf{x}}^{\mathcal{S}}, W_k^{\mathcal{S}} \cdot \overline{\mathbf{x}}^{\mathcal{S}}, W_v^{\mathcal{S}} \cdot \overline{\mathbf{x}}^{\mathcal{S}} \\
Q^{\mathcal{A}}, K^{\mathcal{A}}, V^{\mathcal{A}} &= W_q^{\mathcal{A}} \cdot \overline{\mathbf{x}}^{\mathcal{A}}, W_k^{\mathcal{A}} \cdot \overline{\mathbf{x}}^{\mathcal{A}}, W_v^{\mathcal{A}} \cdot \overline{\mathbf{x}}^{\mathcal{A}}
\end{aligned}
\end{equation}
\vspace{+1pt}
where $W_q^{\cdot}$, $W_k^{\cdot}$, and $W_v^{\cdot}$ denote learnable matrices.

Then we can obtain the intermediate outputs of bidirectional casual cross-attention as:
\begin{equation}
\label{inter}
\begin{aligned}
\mathbf{z}^{\mathcal{S}} &= \text{CCA}(\mathcal{A} \rightarrow \mathcal{S}) = \operatorname{softmax}\left(\frac{Q^{\mathcal{S}} \cdot {K^{\mathcal{A}}}^{\top}}{\sqrt{d^{\mathcal{A}}_k}} + \mathbf{M}\right) \cdot V^{\mathcal{A}}\\
\mathbf{z}^{\mathcal{A}} &= \text{CCA}(\mathcal{S} \rightarrow \mathcal{A}) = \operatorname{softmax}\left(\frac{Q^{\mathcal{A}} \cdot {K^{\mathcal{S}}}^{\top}}{\sqrt{d^{\mathcal{S}}_k}} + \mathbf{M}\right) \cdot V^{\mathcal{S}}
\end{aligned}
\end{equation}
where CCA denotes the casual cross-attention operation, $d^{\cdot}_k$ is the dimension of the corresponding Key matrix, and $\mathbf{M}$ is the causal mask ensuring that the attention mechanism only considers information from past and present time steps.

The intermediate outputs then proceed through the layer normalization and feedforward steps in the transformer architecture, resulting in $\dot{\mathbf{z}}^{\mathcal{S}}$ and $\dot{\mathbf{z}}^{\mathcal{A}}$. These are finally combined using a mean pooling operation to produce a $1 \times T$ dimensional reward sequence output:
\begin{equation}
\mathbf{z}^{\mathcal{R}} = \hat{r}_{1:T} = \mathcal{M}\left(\dot{\mathbf{z}}^{\mathcal{S}}, \dot{\mathbf{z}}^{\mathcal{A}}\right)
\end{equation}
where $\mathcal{M}$ represents the mean pooling operation. 

This step makes PrefMMT capable of capturing the complex interactions between state and action modalities within the trajectories and how these interactions influence human evaluative preferences. For example, in a pick-and-place task, the model can learn that a smooth, arcing motion (action) is preferred when the object (state) is fragile, whereas a direct path may be acceptable for more robust objects.

\subsection{Model Training and Employment}
Overall, the hierarchical outputs of PrefMMT generate a sequence of non-Markovian rewards $\hat{r}_{1:T}$ corresponding to each time step in the trajectory as:
\begin{equation}
\hat{r}_{1:T} = f_{inter}(f_{intra}(f_{e}(s_{1:T}, a_{1:T})))
\end{equation}

We regard the sum of each non-Markovian reward in the sequence as the overall preference score in the Bradley-Terry model in Eq. \ref{BT}, i.e., $\rho_{\mathrm{MMT}} = \sum_{\hat{r}_t \in \mathbf{z}^{\mathcal{R}}} \hat{r}_t$. Note that this sum is not a simple equal sum, as each reward is inherently weighted by a multimodal weight $w^{mul}$ as shown in Fig. \ref{fig:comparision}. These weights are shaped by all the learnable Queries, Keys, and Values in both the intra-modal and inter-modal encoders within PrefMMT, as described in Eqs. \ref{intra}-\ref{inter}. Finally, we train PrefMMT by optimizing the loss function outlined in Eq. \ref{loss}.

To employ PrefMMT as the reward model in subsequent RL training, we consider a sliding window of $T$ past transitions at each time step $t$ to obtain a trajectory input:
$\sigma_t = (\mathbf{s}_{t-T+1}, \mathbf{a}_{t-T+1}, \ldots, \mathbf{s}_t, \mathbf{a}_t)$. Then the reward for the current time step $t$ is taken to be the $t^{th}$ (final) element within the generated reward sequence $\mathbf{z}^{\mathcal{R}}$. This approach ensures that the reward reflects the accumulated preference credit over the relevant sequence of state-action pairs, providing a context-sensitive reward signal that guides the RL agent toward behaviors aligned with human preferences.

\section{Experimental Setups}
\label{case}

In line with previous works \cite{kim2023preference,hejna2023inverse,hejna2024contrastive}, we evaluate PrefMMT in an offline setting: modeling human preferences using offline preference datasets and employing offline RL, specifically Implicit Q-Learning (IQL) \cite{kostrikov2022offline}, for policy learning. This approach ensures a fair comparison by eliminating variations caused by real-time preference feedback and data collection, allowing us to focus solely on the effectiveness of the preference modeling and subsequent policy learning. 

\subsection{Task Environments and Preference Labels}
For experimental environments, we considered three different task domains: AntMaze, Gym-Mujoco locomotion tasks from the D4RL benchmark \cite{fu2020d4rl}, and manipulation tasks from the Meta-World benchmark \cite{yu2020meta}. For preference reward learning, we used preference labels from real humans for the AntMaze and Gym-Mujoco locomotion tasks, as provided by \cite{kim2023preference}, and we collected new preference labels for the manipulation tasks ourselves following the same procedure in \cite{kim2023preference}. For each model, the same set of 100 feedback queries per task was utilized, a practical and realistic volume of human preference data for collection.

\subsection{Baselines and Ablation Studies}
We compared our PrefMMT with three state-of-the-art baselines in PM and one state-of-the-art baseline that directly optimizes policy from preference data:
\begin{itemize}[leftmargin=*]
    \item MR: This represents the most common Markovian reward learning approach with MLP networks \cite{christiano2017deep,lee2021pebble,park2021surf,hiranaka2023primitive,metcalf2023sample,liu2022task,liu2022meta}.
    \item PrefLSTM \cite{early2022non}: This represents a benchmark method that uses Long Short-Term Memory (LSTM) networks to learn non-Markovian rewards.
    \item Preference Transformer (PT) \cite{kim2023preference}: This represents a state-of-the-art sequence modeling baseline that employs unimodal, casual transformers to model non-Markovian rewards and capture temporal dependencies in preferences.
    \item CPL \cite{hejna2024contrastive}: This represents a state-of-the-art preference-based policy learning baseline that bypasses explicit reward modeling by directly using a contrastive objective to align policy behaviors with preferences. The original CPL method involves a supervised pre-training phase, where baseline policies are trained with oracle task rewards until they reach approximately a 50\% success rate, followed by fine-tuning using preference data. For a fair comparison, we removed this supervised RL pre-training step and trained CPL directly from scratch using only preference data, ensuring consistent conditions with the other models.
\end{itemize}

To further evaluate the benefits of each module in our PrefMMT, we also built two ablation models: \begin{itemize}[leftmargin=*]
    \item PrefIntra: Deleting the inter-modal joint encoder in the PrefMMT, using mean pooling to generate reward sequences with the outputs from intra-modal encoders.
    \item PrefInter: Removing the intra-modal encoders in the PrefMMT, directly feeding state and action sequences into the inter-modal joint encoder.
\end{itemize}

\subsection{Evaluation}
For evaluation metrics, we utilized the expert-normalized scores: $100\times\frac{\text{score}-\text{random score}}{\text{expert score}-\text{random score}}$, as introduced in the D4RL benchmark \cite{fu2020d4rl} for AntMaze and locomotion tasks, and used success rate for manipulation tasks on the Meta-World benchmark. For each task, we conducted five independent runs for each model and reported the mean and standard deviation. We also reported the performance of IQL with task rewards as an oracle baseline. More implementation details are available on our website.

\begin{table*}[t]
\centering
\caption{Comparison of PrefMMT with baselines and ablation models in terms of normalized scores on Gym and AntMaze tasks from D4RL, and success rates on manipulation tasks from Meta-World using 100 human preference queries. Mean and standard deviation are reported over five independent runs.  Oracle performance of IQL with task reward is also reported. The highest performance on each task is highlighted.}
\small
\newcommand{\highlight}[1]{\colorbox{orange!15}{#1}}
\newcommand{\avghighlight}[1]{\colorbox{red!15}{#1}}
\resizebox{\linewidth}{!}{
\begin{tabular}{c|c|cccc|cc|c}
\toprule
\multirow{2}{*}{\parbox{1cm}{\centering \textit{Task}}} & \multirow{2}{*}{\parbox{1.2cm}{\centering \textit{IQL with task reward}}} & \multicolumn{7}{c}{\textit{Preference Learning}} \\ 
\cmidrule{3-9}
& & \textit{MR} & \textit{PrefLSTM} & \textit{PT} & \textit{CPL} & \textit{PrefIntra}& \textit{PrefInter} & \textit{PrefMMT} \\ 
\midrule
walker2d-medium-expert-v2 & 103.31 {\scriptsize $\pm$ 1.18} & 99.77 {\scriptsize $\pm$ 2.31} & 93.96 {\scriptsize $\pm$ 6.90} & 103.52 {\scriptsize $\pm$ 0.45} &93.70 {\scriptsize $\pm$ 0.49} & 102.56 {\scriptsize $\pm$ 1.43} & 104.97 {\scriptsize $\pm$ 1.24} & \highlight{113.00 {\scriptsize $\pm$ 0.57}} \\ 
walker2d-medium-replay-v2 & 73.03 {\scriptsize $\pm$ 0.79} & 71.47 {\scriptsize $\pm$ 3.81} & 63.02 {\scriptsize $\pm$ 6.27} & \highlight{75.48 {\scriptsize $\pm$ 1.53}} &43.67 {\scriptsize $\pm$ 3.76} & 60.81 {\scriptsize $\pm$ 13.60} & 65.96 {\scriptsize $\pm$ 8.58} & 75.35 {\scriptsize $\pm$ 0.21} \\ 
hopper-medium-expert-v2 & 69.20 {\scriptsize $\pm$ 3.23} & 76.91 {\scriptsize $\pm$ 1.04} & 59.41 {\scriptsize $\pm$ 7.75} & \highlight{83.77 {\scriptsize $\pm$ 3.01}} &66.42{\scriptsize $\pm$ 0.98} & 68.53 {\scriptsize $\pm$ 6.58} & 78.65 {\scriptsize $\pm$ 1.30} & 80.27 {\scriptsize $\pm$ 6.21} \\ 
hopper-medium-replay-v2 & 58.25 {\scriptsize $\pm$ 17.35} & 29.73 {\scriptsize $\pm$ 3.98} & 51.12 {\scriptsize $\pm$ 8.23} & 69.24 {\scriptsize $\pm$ 0.15} &83.63 {\scriptsize $\pm$ 9.97}& 53.45 {\scriptsize $\pm$ 14.86} & 64.18 {\scriptsize $\pm$ 17.27} & \highlight{84.40 {\scriptsize $\pm$ 2.45}} \\ 
\midrule
\textbf{Gym-Average} & 75.95 {\scriptsize $\pm$ 5.64} & 69.47 {\scriptsize $\pm$ 2.79} & 66.88 {\scriptsize $\pm$ 7.29} & 83.00  {\scriptsize $\pm$ 1.29} &71.86 {\scriptsize $\pm$ 3.80}& 71.34 {\scriptsize $\pm$ 9.12} & 78.44 {\scriptsize $\pm$ 7.10} & \avghighlight{88.26 {\scriptsize $\pm$ 2.36}} \\ 
\midrule
Antmaze-large-play-v2 & 35.55 {\scriptsize $\pm$ 1.75} & 8.10 {\scriptsize $\pm$ 1.10} & 6.71 {\scriptsize $\pm$ 4.66} & 18.20  {\scriptsize $\pm$ 1.60} & 15.39 {\scriptsize $\pm$ 2.65} & 16.75 {\scriptsize $\pm$ 0.75} & 18.45 {\scriptsize $\pm$ 7.25} & \highlight{41.03 {\scriptsize $\pm$ 1.41}} \\ 
Antmaze-large-diverse-v2 & 32.20 {\scriptsize $\pm$ 2.10} & 1.66 {\scriptsize $\pm$ 0.65} & 0.00 {\scriptsize $\pm$ 0.00} & 16.65  {\scriptsize $\pm$ 3.65} & 10.27 {\scriptsize $\pm$ 1.86} & 11.55 {\scriptsize $\pm$ 3.15} & 11.1 {\scriptsize $\pm$ 0.20} & \highlight{38.58 {\scriptsize $\pm$ 5.19}} \\
Antmaze-medium-play-v2 & 67.35 {\scriptsize $\pm$ 1.25} & 48.75 {\scriptsize $\pm$ 3.95} & 15.41 {\scriptsize $\pm$ 5.21} & \highlight{67.05 {\scriptsize $\pm$ 1.45}} & 45.18 {\scriptsize $\pm$ 3.57}  & 61.55 {\scriptsize $\pm$ 0.55} & 62.00 {\scriptsize $\pm$ 1.40} & 66.21 {\scriptsize $\pm$ 2.45} \\ 
Antmaze-medium-diverse-v2 & 60.00 {\scriptsize $\pm$ 1.90} & 10.70 {\scriptsize $\pm$ 0.20} & 15.55 {\scriptsize $\pm$ 5.55} & 62.40 {\scriptsize $\pm$ 2.80} & 41.63 {\scriptsize $\pm$ 3.41}  & 61.40 {\scriptsize $\pm$ 0.70} & 59.95 {\scriptsize $\pm$ 2.35} & \highlight{79.85 {\scriptsize $\pm$ 0.94}} \\ 
\midrule
\textbf{AntMaze-Average} & 48.78 {\scriptsize $\pm$ 1.75} & 17.30 {\scriptsize $\pm$ 1.48} & 9.42 {\scriptsize $\pm$ 3.86} & 41.08 {\scriptsize $\pm$ 2.38} & 28.12 {\scriptsize $\pm$ 2.87} & 37.81 {\scriptsize $\pm$ 1.29} & 37.88 {\scriptsize $\pm$ 2.80} & \avghighlight{56.42 {\scriptsize $\pm$ 2.50}} \\ 
\midrule
Sweep Into & 43.52 {\scriptsize $\pm$ 0.67} & 46.68 {\scriptsize $\pm$ 1.22} & 45.65 {\scriptsize $\pm$ 0.35} & 44.70 {\scriptsize $\pm$ 1.20} &22.75{\scriptsize $\pm$2.62} & 45.35 {\scriptsize $\pm$ 1.40} & 43.98 {\scriptsize $\pm$ 1.58} & \highlight{58.17 {\scriptsize $\pm$ 1.41}} \\ 
Drawer Open & 66.43 {\scriptsize $\pm$ 0.98} & 64.71 {\scriptsize $\pm$ 1.59} & 64.90 {\scriptsize $\pm$ 0.95} & 65.55 {\scriptsize $\pm$ 0.7} &34.91{\scriptsize $\pm$3.84} & 65.45 {\scriptsize $\pm$ 1.95} & 66.22 {\scriptsize $\pm$ 1.12} & \highlight{81.21 {\scriptsize $\pm$ 3.56}} \\ 
Button Press & 67.82 {\scriptsize $\pm$ 0.42} & 66.38 {\scriptsize $\pm$ 0.88} & 68.80 {\scriptsize $\pm$ 0.15} & 67.65 {\scriptsize $\pm$ 1.65} &44.76{\scriptsize $\pm$3.61} & 68.6 {\scriptsize $\pm$ 1.65} & 67.3 {\scriptsize $\pm$ 1.85} & \highlight{78.64 {\scriptsize $\pm$ 1.41}} \\
Window Close & 72.42 {\scriptsize $\pm$ 0.53} & \highlight{75.40 {\scriptsize $\pm$ 1.15}} & 70.42 {\scriptsize $\pm$ 1.77} & 71.47 {\scriptsize $\pm$ 2.12} &48.13{\scriptsize $\pm$5.37} & 70.74 {\scriptsize $\pm$ 1.16} & 71.22 {\scriptsize $\pm$ 1.28} & 74.40 {\scriptsize $\pm$ 2.05} \\ 
\midrule
\textbf{MetaWorld-Average} & 62.55 {\scriptsize $\pm$ 0.65} & 63.29 {\scriptsize $\pm$ 1.21} & 62.44 {\scriptsize $\pm$ 0.81} & 62.34 {\scriptsize $\pm$ 1.42} &37.64{\scriptsize $\pm$ 3.86} & 62.54 {\scriptsize $\pm$ 1.54} & 62.18 {\scriptsize $\pm$ 1.46} & \avghighlight{73.11 {\scriptsize $\pm$ 2.11}} \\ 
\bottomrule 
\end{tabular}}
\label{tab:IQL_result}
\end{table*}

\section{Results and Analysis}
\subsection{Quantitative Measurement}
Table \ref{tab:IQL_result} presents the performance comparisons of different reward modeling methods across various tasks. Our PrefMMT outperforms all baselines in terms of average performance across all task domains and leads on 8 out of 12 sub-tasks. Notably, PrefMMT even surpasses the oracle performance of IQL with task reward in the Gym locomotion and Meta-World domains. These results demonstrate that PrefMMT can serve as a more robust and effective preference modeling approach, capable of inferring meaningful reward patterns from real human preferences and leading to efficient robot behaviors.

\subsubsection{Comparison with PM Baselines}
Compared to MR, while both PT and PrefMMT show improvement, another non-Markovian baseline PrefLSTM does not demonstrate better performance, even failing on the AntMaze-large-diverse task, which aligns with findings in \cite{kim2023preference}. We attribute this to the limitations of LSTM-based models in capturing complex, long-range dependencies within trajectories, which are crucial for accurately modeling human preferences in challenging environments.

While PT demonstrates competitive performance, it exhibits a larger standard deviation and struggles in more complex or larger task environments, such as the large/diverse settings of AntMaze and Sweep Into. We believe this performance difference arises because, although PT can capture temporal dependencies and infer critical events, such as identifying key waypoints in AntMaze and detecting object contact in Sweep Into, PrefMMT not only covers these aspects but also models cross-modal interactions between states and actions.

For instance, in AntMaze, PrefMMT can recognize how specific leg movements (actions) correspond to changes in the agent's orientation and velocity (states), enabling it to prefer trajectories where the ant maintains balance while making efficient progress. Similarly, in Sweep Into, PrefMMT can capture the relationship between the end-effector and target positions (states) and the force applied (actions) during the sweeping motion. This allows PrefMMT to prefer trajectories where the robot adjusts its force based on the object's position and movement, leading to smoother and more controlled sweeping actions. These insights highlight the advantages of modeling the multimodal nature of robot trajectories and capturing state-action interactions for preference modeling.

On the other hand, while PrefMMT does not achieve the highest performance on 4 out of 12 sub-tasks, it demonstrates a higher correlation with real human preferences on these specific tasks. To validate this, we compared the Pearson correlation \cite{sedgwick2012pearson} between real human preference labels and synthetic preference labels generated by PrefMMT and the baselines that outperformed it, specifically MR on Window Close and PT on AntMaze and Gym sub-tasks, across an additional 50 pairs of unseen trajectories.

\begin{figure*}[t]
\centering
\includegraphics[width=\linewidth]{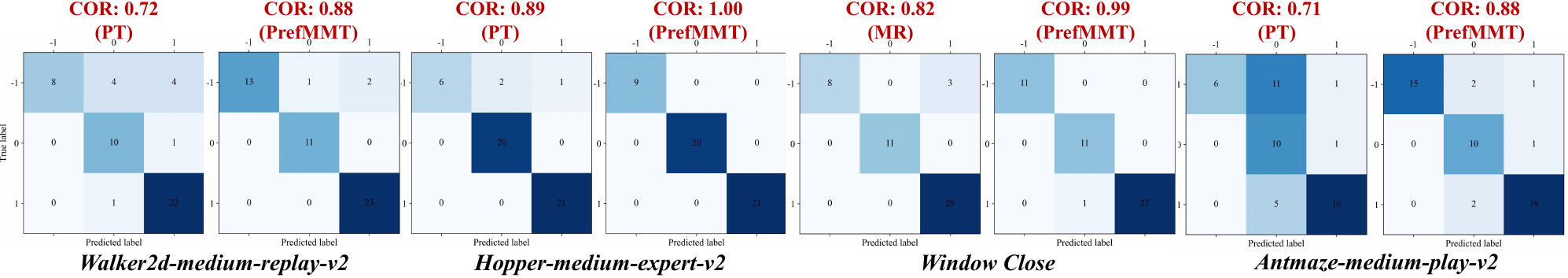}
\vspace{-10pt}
\caption{Confusion matrices and Pearson correlation (COR) of real human preference labels (y-axis) and predicted preference labels from different PM models. Labels: 1 and 0 denote a preference for the first or second trajectory, respectively, while -1 indicates indecision.}
\vspace{-10pt}
\label{fig:cor}
\end{figure*}

As shown in Fig. \ref{fig:cor}, PrefMMT consistently achieves a higher correlation, aligning more closely with real human preferences. This finding suggests that PrefMMT holds greater potential for preference-driven tasks, particularly in human-robot interaction scenarios where alignment with human preferences is often more critical than general task performance.

\subsubsection{Comparison with Direct Preference-Based Policy Learning Baselines}
Furthermore, we observe that our method consistently outperforms CPL, particularly in more complex tasks, such as AntMaze, Sweep-into on MetaWorld, and Walker-replay in locomotion. Interestingly, while \cite{hejna2024contrastive} reports significant performance gains, CPL performs worse in our setting. We attribute this discrepancy to two unrealistic assumptions in the original experimental setting of the CPL.

One key assumption in CPL is the need for supervised pre-training by training baseline policies with oracle task rewards until they approximately reach a 50\% success rate. This requirement may not hold in scenarios where task rewards are unavailable or unreliable, limiting CPL's applicability in settings that rely solely on preference data.

Another issue lies in the reliance on an oracle policy trained with Soft Actor-Critic (SAC) to a 100\% success rate to generate synthetic regret-based preference labels. These synthetic labels are not only \textit{idealized} but also \textit{abundant}, with thousands of labels providing a dense and highly accurate preference dataset. In contrast, our experiments use real human feedback, limited to only 100 preference queries, where human preferences can be \textit{noisy} and exhibit \textit{uncertainty}. This discrepancy highlights that while CPL performs well under ideal conditions with ample, perfectly labeled data, our approach is more robust in realistic settings with limited and potentially noisy human feedback.

These findings align with \cite{nika2024reward, xudpo}, which demonstrate the potential limitations in sample efficiency and stability when directly learning policies from preference data. Together, these insights emphasize that improving the PM remains a promising direction, especially for scenarios where only preference data is available, and the amount of human feedback is more realistic. Our approach reinforces the value of PM by explicitly modeling the multimodal nature of robot trajectories, capturing both intra- and inter-modal dynamics to better align with human preferences.

\subsubsection{Ablation Results}

Additionally, compared to the ablation models, PrefIntra and PrefInter, PrefMMT demonstrates superior performance across all tasks. While these ablation models incorporate modality decoupling, they do not effectively capture both intra-modal temporal dependencies and inter-modal state-action interactions.

Specifically, PrefIntra may struggle with state-action cross-modal reasoning, such as correlating an object's position (state) with the robot's sweeping motion (action). Conversely, PrefInter may fall short in capturing long-term temporal dependencies, such as planning a sequence of sweeps to efficiently clear an entire area.

These results highlight the importance of PrefMMT's hierarchical transformer architecture, which enables comprehensive modeling of complex intra- and inter-modal relationships, leading to more nuanced and accurate preference modeling.

\begin{figure*}[t]
\centering
\includegraphics[width=\linewidth]{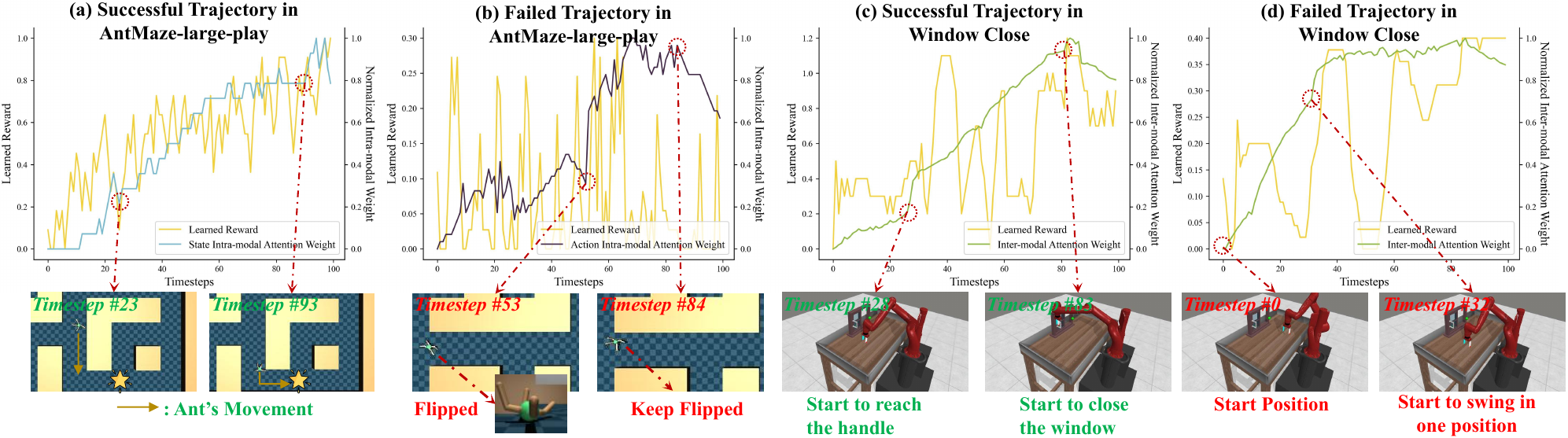}
\vspace{-10pt}
\caption{Series of learned preference rewards (yellow) along with normalized state (cyan) and action (purple) intra-modal attention weights, and state-action inter-modal (green) attention weights from PrefMMT on successful and failed trajectories in the AntMaze-large-play-v2 and Window Close tasks. Stars present the escape goals in AntMaze (the figure supports zooming in for more detailed information).}
\vspace{-5pt}
\label{fig:visual}
\end{figure*}

\subsection{Qualitative Analysis of Preference Reward and Attention}
To further investigate the preference attention mechanism within PrefMMT, we visualize the learned reward values along with the normalized intra-modal attention weights of the state and action modalities, and the normalized state-action inter-modal attention weights. Fig. \ref{fig:visual} shows these visualizations for both successful and failed robot trajectories on the AntMaze-large-play-v2 and Window Close tasks. A zoomed and GIF version can be found at the project website. 

We observe that the rewards generated by PrefMMT align well with human expectations and task requirements. PrefMMT assigns higher rewards to promising behaviors in successful trajectories with an approximate maximum of 1.2, such as approaching the escape goal (Fig. \ref{fig:visual}a) or approaching and closing the window (Fig. \ref{fig:visual}c). Conversely, PrefMMT effectively penalizes ineffective behaviors in failed trajectories with an approximate maximum of 0.35, such as getting trapped in the maze (Fig. \ref{fig:visual}b) or exhibiting a swinging motion without progress (Fig. \ref{fig:visual}d).

We also find that the preference attention learned in both the intra-modal and inter-modal joint encoders is meaningful. In Fig. \ref{fig:visual}a, we observe that the state intra-modal attention weight increases significantly at time step 23 when the ant changes direction and approaches the escape goal, and at step 93 when it is about to turn the corner and reach the goal (positive state changes). Moreover, in Fig. \ref{fig:visual}b, the action attention weight rises sharply at step 53 when the ant flips over, and remains high until step 84 as it continues to struggle on the ground (negative action sequences while the state, ant position, remains almost unchanged). These examples demonstrate that the intra-modal attention in PrefMMT effectively captures important temporal dependencies in state transitions and action sequences that influence human judgments in both successful and failed trajectories.

Moreover, we observe in Fig. \ref{fig:visual}c that that the state-action inter-modal attention weight increases significantly at step 28 when the robot arm starts reaching for the window handle, and again at step 83 when it begins to close the window by pushing the handle, remaining high during and after this process as the arm continues reaching and closing. Similarly, in Fig. \ref{fig:visual}d, the attention weight rises between steps 0 and 32 while the robot arm attempts to reach the handle, and increases further after step 32 when the arm starts swinging aimlessly at its current position, stuck. These findings demonstrate that the inter-modal attention in PrefMMT effectively captures state-action interplays that influence human preferences, whether positively or negatively.

\section{Conclusion}
In this paper, we propose PrefMMT, addressing the preference modeling problem in PbRL by treating it as a multimodal sequence modeling task that accounts for the multimodal nature of robot trajectories. We propose a hierarchical multimodal transformer network that models temporal dependencies within the state and action modalities while capturing the inter-modal interactions that influence human judgments. Extensive experiments on RL benchmarks demonstrate the benefits of PrefMMT, showing consistent performance gains over state-of-the-art PM and direct policy learning baselines. Our results highlight the importance of explicitly modeling both intra- and inter-modal dynamics for robust and accurate preference modeling, especially in complex and realistic scenarios with limited human feedback.

\typeout{}
\bibliography{main}
\bibliographystyle{IEEEtran}

\end{document}